\pdfoutput=1

\documentclass[11pt]{article}

\usepackage[final]{acl}

\usepackage{times}
\usepackage{latexsym}

\usepackage[T1]{fontenc}

\usepackage[utf8]{inputenc}

\usepackage{microtype}

\usepackage{inconsolata}

\usepackage{graphicx}

\title{Psycholinguistic Word Features: a New Approach for the Evaluation of LLMs Alignment with Humans}

\author{Javier Conde, Miguel Gonz\'alez, Mar\'ia Grandury, \\
\textbf{Gonzalo Martínez, Pedro Reviriego} \\
Universidad Polit\'ecnica de Madrid \\
Madrid, Spain \\
\And
 Marc Brysbaert \\
Ghent University \\
Ghent, Belgium \\
}

\begin{document}
\maketitle
\begin{abstract}
The evaluation of LLMs has so far focused primarily on how well they can perform different tasks such as reasoning, question-answering, paraphrasing, or translating. For most of these tasks, performance can be measured with objective metrics, such as the number of correct answers. However, other language features are not easily quantified. For example, arousal, concreteness, or gender associated with a given word, as well as the extent to which we experience words with senses and relate them to a specific sense. Those features have been studied for many years by psycholinguistics, conducting large-scale experiments with humans to produce ratings for thousands of words. This opens an opportunity to evaluate how well LLMs align with human ratings on these word features, taking advantage of existing studies that cover many different language features in a large number of words. 

In this paper, we evaluate the alignment of a representative group of LLMs with human ratings on two psycholinguistic datasets: the Glasgow and Lancaster norms. These datasets cover thirteen features over thousands of words. The results show that alignment is \textcolor{black}{generally} better in the Glasgow norms evaluated (arousal, valence, dominance, concreteness, imageability, familiarity, and gender) than on the Lancaster norms evaluated (introceptive, gustatory, olfactory, haptic, auditory, and visual). This suggests a \textcolor{black}{potential} limitation of current LLMs in aligning with human sensory associations for words, which may be due to their lack of embodied cognition present in humans and illustrates the usefulness of evaluating LLMs with psycholinguistic datasets.
\end{abstract}

\section{Introduction}

The evaluation of Large Language Models (LLMs) poses significant challenges as they have to be evaluated on their performance on a large number of tasks and their answers are in natural language \cite{LLM_evaluation_survey}. One alternative is to have humans evaluate the LLM responses. This, however, does not scale when an extensive evaluation with tens of thousands of questions has to be done for each model and new models appear every day. Initiatives like the Chatbot Arena \cite{chatbot_arena} resort to the community to perform an evaluation of human preferences. In this case, the questions, answers, and participants are not controlled, so the results provide a comparative ranking of models but not a detailed analysis of their specific capabilities. Another alternative is to use an LLM to evaluate other LLMs \cite{llmjudging}. Again, this method has limitations as the judging LLM may have biases and inaccuracies, and someone has to evaluate this LLM in the first place. The most widely used method to evaluate LLMs as of today is to run different benchmarks, mostly made of multiple-choice questions or tasks for which existing metrics can be used to provide a result. This enables the automation of the process and the evaluation of specific tasks, for example, maths \cite{Mathmeasuring}, reasoning \cite{CommonSensemeasuring}, or knowledge of many different topics \cite{MMLU, BIGMeasuring}. The results of those tests are then published on leaderboards \cite{open-llm-leaderboard-v2, LeaderbaordOpenQuestions} and used to compare the performance of LLMs on a wide range of tasks.

Evaluating LLMs' ability to solve a math problem, a riddle, or answer a question about taxation is interesting but is not enough. LLMs interact with persons and generate text that is read by humans. Therefore, we would like them to be aligned with human emotions, perceptions and preferences \cite{Alignment1, Alignment2}. To assess alignment, benchmarks for emotional alignment are also being developed, for example, by asking open questions to the LLM and evaluating their responses using a second LLM as a judge \cite{emotioneval1}. This, as discussed before, relies on the judge LLM and thus is limited by its capabilities. Another option is to have humans rate the questions on a Likert scale and then ask the LLMs to also answer on a Likert scale \cite{emotioneval2}. This requires new human studies which imply a significant effort \cite{emotioneval2}. Interestingly, human ratings have been used in psycholinguistics for decades and large datasets are available, for example, with ratings of words and expressions \cite{warriner2013norms}. 

Although LLMs are entirely based on written language, they capture much of the meaning of words. For example, LLM-based estimates of the valence and concreteness of words and expressions correlate very well with human ratings \cite{trott2024can,martinez2025using}. At the same time, it is hard to deny that for humans word meaning is more than the occurrence of words together, which is from what LLMs learn. Two aspects come into play here. The first is the symbol grounding problem \cite{harnad1990symbol}. You cannot learn a language on the basis of words alone. Some words must first be grounded in the world around us (at least 1\% according to \cite{vincent2016latent}, or about 400 words). Only then can they be used to accurately define the meaning of other words. The second aspect is that even though words can be defined from other words, in reality we have probably learned their full meaning through a mix of language and everyday experiences. The latter includes perception (our knowledge of the color purple is more than knowing it is a combination of red and blue), actions (our knowledge of a chair is based in part on having sat on chairs many times), emotions, social interactions, and so on. \textcolor{black}{Finally, theories like embodied cognition argue that the interactions of our body with the environment also shape our minds and are an essential part of our language learning process and influence word meaning \cite{wilson2002six}, \cite{barsalou2008grounded}}. Therefore, it is interesting to study whether these fundamental differences between humans and LLMs limit their alignment and in which areas.  

In psycholinguistics, ratings of words and expressions are used to select stimuli for experiments that evaluate different aspects of language processing and learning, supporting the development and validation of theories of human cognition \cite{ExperimentsPsycholinguistics}. Features such as arousal, valence, concreteness, dominance and iconicity have been evaluated on thousands of words and expressions in many different languages \cite{Psycholinguistic_features_databases}. There are also studies with human ratings on different emotions such as happiness, disgust, anger, fear, or sadness \cite{HinojosaEmotions} which are useful in affective neurolinguistics studies \cite{ANL1}. Ratings of how humans associate words with the senses or parts of the body are also available for thousands of words \cite{lynott2020lancaster} \textcolor{black}{and have been used to enrich language models \cite{kennington-2021-enriching}}. Since all these datasets are available and have been used and validated in many studies, it is of interest to explore whether they can be used to evaluate LLMs. \textcolor{black}{So, differently from existing studies \cite{trott2024can,martinez2025using} that use LLMs to generate estimates of word features, use existing human ratings to evaluate LLMs.}

In this paper, we make the first contribution in this direction by presenting an initial study on the use of psycholinguistic datasets for LLM evaluation and analyzing the results linking them to existing works in cognitive science. The rest of the paper is organized as follows. Section \ref{sec:objectives} presents the motivation and objectives of the paper. Section \ref{sec:methodology} presents the evaluation methodology including the selection of the datasets, the LLMs to evaluate and the procedures and metrics used. The results are presented in section \ref{sec:results} and discussed in section \ref{sec:discussion}. The paper ends with the conclusion in section \ref{sec:conclusion}.

\section{Motivation and objectives}
\label{sec:objectives}

The main motivation of this work is to foster the evaluation of LLMs from a psycholinguistic perspective, reusing existing datasets and knowledge that have been gathered in human evaluations for decades. This would not only provide datasets for LLM evaluation but also open new perspectives on how to evaluate LLMs and attract the psycholinguist community to LLM evaluation research \cite{borghi2024language}. For example, theories of language acquisition and processing that have been developed for humans can be used to better understand how LLMs process language. 

To achieve this main goal, in this paper we conduct an initial exploration to show the potential of putting together psycholinguistic word norms and LLM evaluation with the following objectives:

\begin{itemize}
    \item Propose a methodology to evaluate the alignment of LLMs and humans using word norms.
    \item Conduct an initial evaluation using a relevant set of word norms and LLMs.
    \item Analyze the results and link them to existing results in psycholinguistics and cognitive science.
    \item Discuss avenues to continue this work. 
\end{itemize}

The following sections address each of these objectives in detail.  

\section{Methodology}
\label{sec:methodology}

This section discusses the proposed methodology to evaluate the alignment of LLM with humans using psycholinguistic word norms. The methodology includes the selection of psycholinguistic datasets, LLM, and the metrics and procedures used in the evaluation. 

\subsection{Datasets}

To have a comprehensive evaluation, as many word norms as possible should be evaluated covering different aspects of word meaning. The norms should cover a significant number of words and ideally be available in several languages. Unfortunately, there is no such psycholinguistic dataset, and the information is spread among different studies, each covering only a set of norms and typically one or at most a few languages. Therefore, the first step is to select a group of existing word norms for evaluation.

For this initial study, we have selected two datasets: 

\begin{itemize}
    \item \textit{The Glasgow norms \cite{scott2019glasgow}} provide human ratings on arousal, valence, dominance, concreteness, imageability, familiarity and gender association for 5,553 English words.
    \item \textit{The Lancaster norms \cite{lynott2020lancaster}} provide human ratings on 1) six perceptual modalities associated with words, touch, hearing, smell, taste, vision, and interoception and 2) on five parts of the body associated with words, mouth/throat, hand/arm, foot/leg, head excluding mouth/throat, and torso. Both for 39,707 English words.
\end{itemize}

The ratings of the body parts associated with words in the Lancaster norms are not used in our evaluation because the instructions given to humans include images showing the body parts that can only be provided to multimodal models and most of the models evaluated are pure LLMs. Therefore, a total of seven word features from the Glasgow norms and six perceptual modalities are used in our study.

The rationale for our selection is that the two datasets cover a relevant number of norms and words in English, which is the dominant language for LLM design and optimization. The Glasgow norms focus on features for which previous works have shown good alignment of leading LLMs such as GPT-4 \cite{trott2024can,martinez2025using}. Therefore, it is of interest to see if this alignment also occurs for other less powerful LLMs. The Lancaster norms instead focus on perceptual norms, which are expected to correlate less with LLMs which lack embodied cognition. 

\subsection{LLMs}

In order to ensure that the results are representative of the current LLMs, we select several open models such as Llama-3.2-3B, LLama3.1-8B \cite{llama3_1}, LLama3.2-11B from Meta AI, Gemma-2-9B \cite{team2024gemma2} from Google, two models optimized for languages other than English: Yi-1.5-9B \cite{ai2024yi} and Occiglot-7B \cite{avramidis2024occiglot} and two proprietary models, OpenAI's GPT-4o and GPT-4o-mini \cite{openai2023gpt4}. As with the datasets, the selection is intended to provide good coverage of the current LLM ecosystem while keeping the computational effort manageable. On one hand, several models with different sizes are evaluated for LLama and GPT-4o to assess the impact of model size. Additionally, for LLama, a multimodal model (LLama3.2-11B ) is included in the evaluation to see if multimodality has any impact on alignment. On the other hand, models from three different companies are evaluated to see if the alignment changes significantly across model families. 


\subsection{Procedure}

We ask the LLMs to rate the words on the different features using as prompts the same questions used in the human studies, adding a sentence to request the LLM to answer only with the number of the rating for the word. This is consistent with previous studies on generating psycholinguistic data with LLMs on which these prompts achieved good results. Two examples of prompts are given below:

\begin{itemize}
    \item \textit{Prompt for Arousal (Glasgow norms):} Arousal is a measure of excitement versus calmness. A word is AROUSING if it makes you feel stimulated, excited, frenzied, jittery, or wide-awake. A word is UNAROUSING if it makes you feel relaxed, calm, sluggish, dull, or sleepy. Please indicate how arousing you think word “X” is on a scale of 1 (VERY UNAROUSING) to 9 (VERY AROUSING), with the midpoint representing moderate arousal. Please answer only with the number.
    \item \textit{Prompt for Gustatory (Lancaster norms):} You will be asked to rate how much you experience everyday concepts using perceptual senses. There are no right or wrong answers so please use your own judgement. The rating scale runs from 0 (not experienced at all with that sense) to 5 (experienced greatly with that sense). Please answer only with the number. To what extent do you experience by tasting word “X” 
\end{itemize}

The temperature of the LLM is set to zero to ensure that results are reproducible and two estimates are computed. The first is the direct answer of the LLM which corresponds to the number with the largest estimated probability. The second estimate is computed by obtaining the LLM estimated probabilities \cite{logprobs} of each of the possible values on the rating scale (typically 0-5, 1-7 or 1-9), multiplying the values by their probabilities and adding them; thus taking the average value given by the estimated probabilities. This second estimate has been shown to be better in previous studies \cite{logprobs}.  

\subsection{Metrics}

To measure the alignment of LLMs with humans, it seems natural to use the metrics that are used in psycholinguistics to check the agreement of different studies that collect ratings on the same word features. Two single value metrics \cite{PearsonSpearman} are commonly used:

\begin{itemize}
    \item \textit{Pearson correlation coefficient:} the covariance of the variables divided by the product of their standard deviations.  
    \item \textit{Spearman correlation coefficient:} the Pearson's correlation of rank variables rather than variables themselves, so it focuses on monotonic relations rather than linear relations. 
\end{itemize}

Pearson correlation coefficient assumes a normal distribution and mainly weighs observations far away from the mean. Spearman correlation coefficient gives equal weight to the entire distribution and may therefore emphasize small differences around the mode. These are important differences because for some of the perceptual norms, the values of both humans and LLMs are concentrated at the lower end of the range (e.g., only a few words are related to smell or touch). To address these issues, we will compute both coefficients on both the original data and values rounded to the nearest integer. The latter agrees more with human experience, as the difference between Likert values of 1.01 and 1.02 is not psychologically meaningful (both values indicate that the words are barely related to characteristic tested).

All in all, four values will be computed:



\begin{itemize}
    \item Pearson coefficient on original human data and the logprob-based estimate for LLMs.
    \item Pearson coefficient on the two metrics above rounded to the nearest integer.
    \item Spearman coefficient on original human data and the logprob-based estimate for LLMs.
    \item Spearman coefficient on the two metrics above rounded to the nearest integer.
\end{itemize}

\section{Results}
\label{sec:results}


\textcolor{black}{All results and prompts used as well as the code to generate the plots are available in a public repository\footnote{\url{https://zenodo.org/records/14866800}}}. The results for the Glasgow	
norms are presented first. As discussed in the previous section, in the following, only the estimate based on the LLM estimated probabilities is used to present the results as, in general, it achieves better alignment with humans. 

\begin{figure*}[t!]
  \centering
  \includegraphics[width=0.47\textwidth]{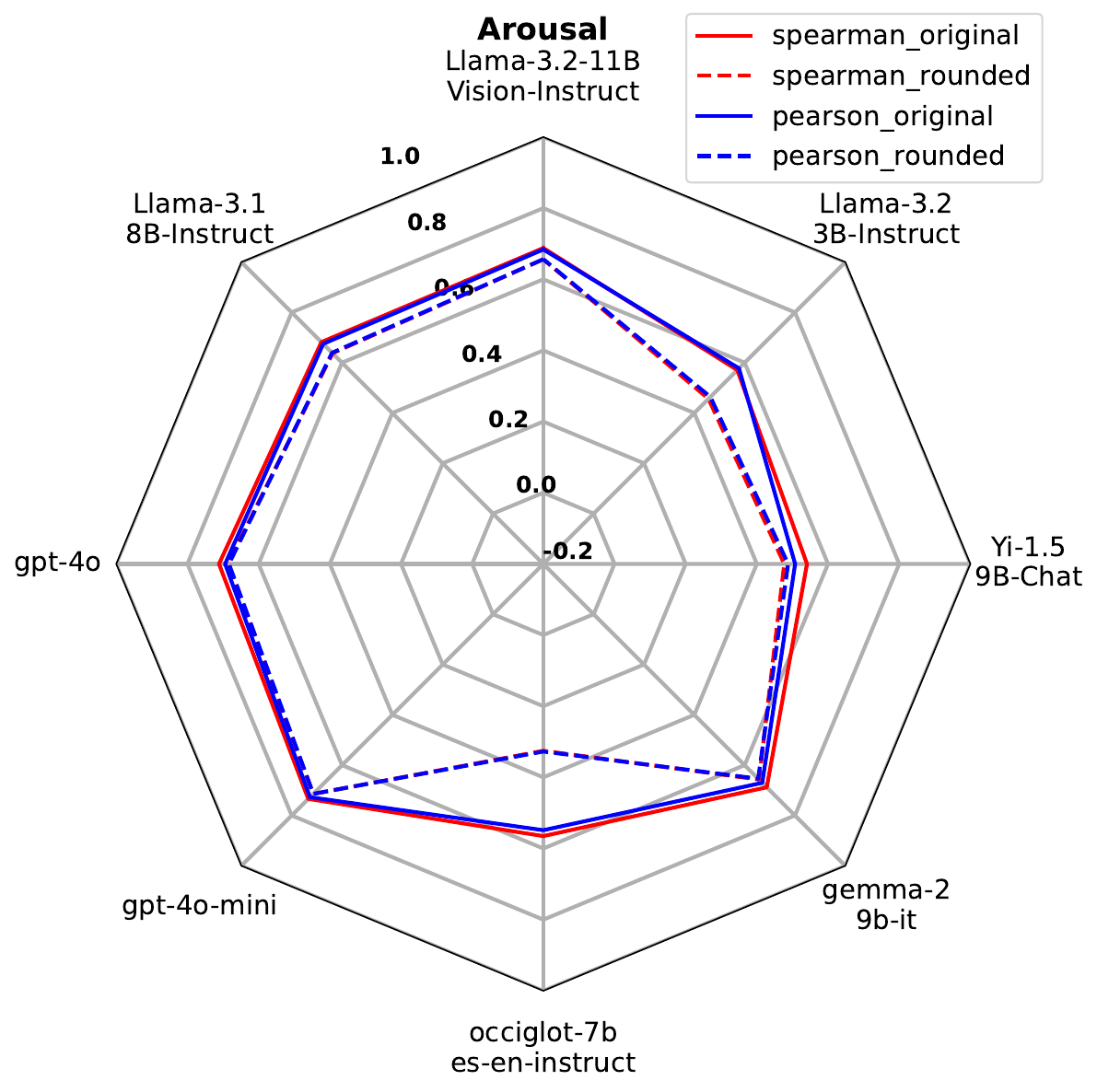}
  \hfill
  \includegraphics[width=0.47\textwidth]{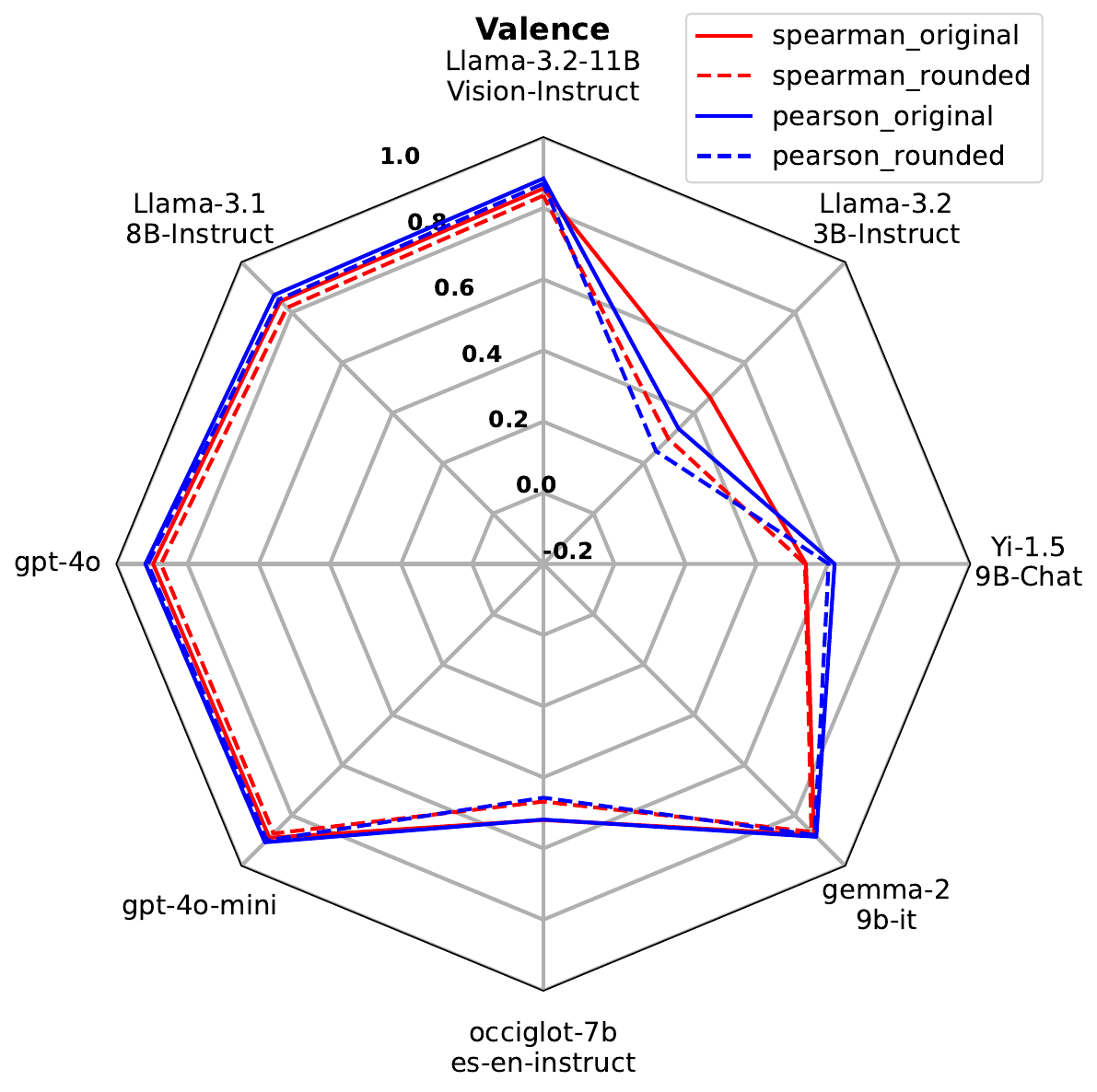}
  \includegraphics[width=0.47\textwidth]{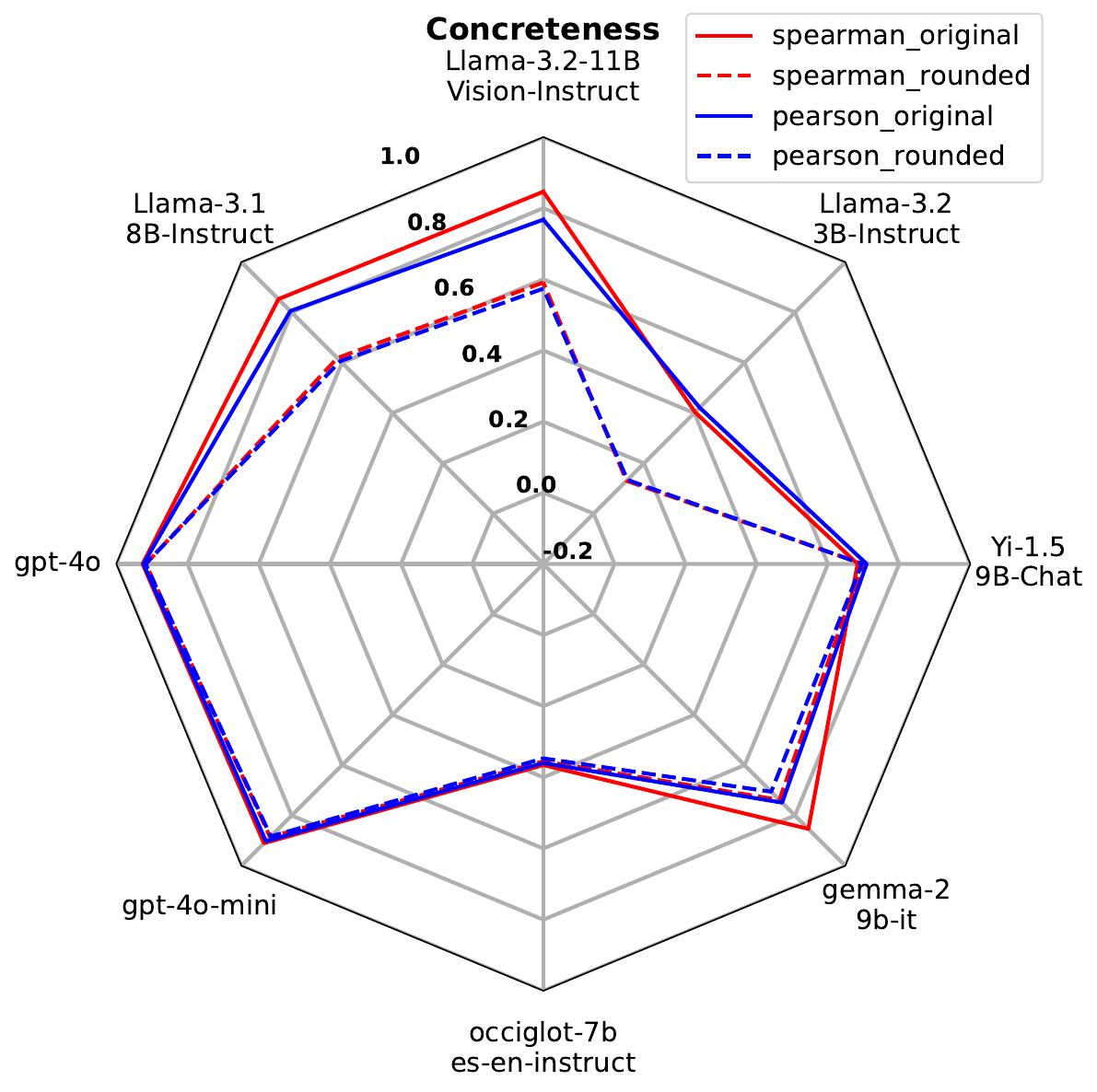}
  \hfill
  \includegraphics[width=0.47\textwidth]{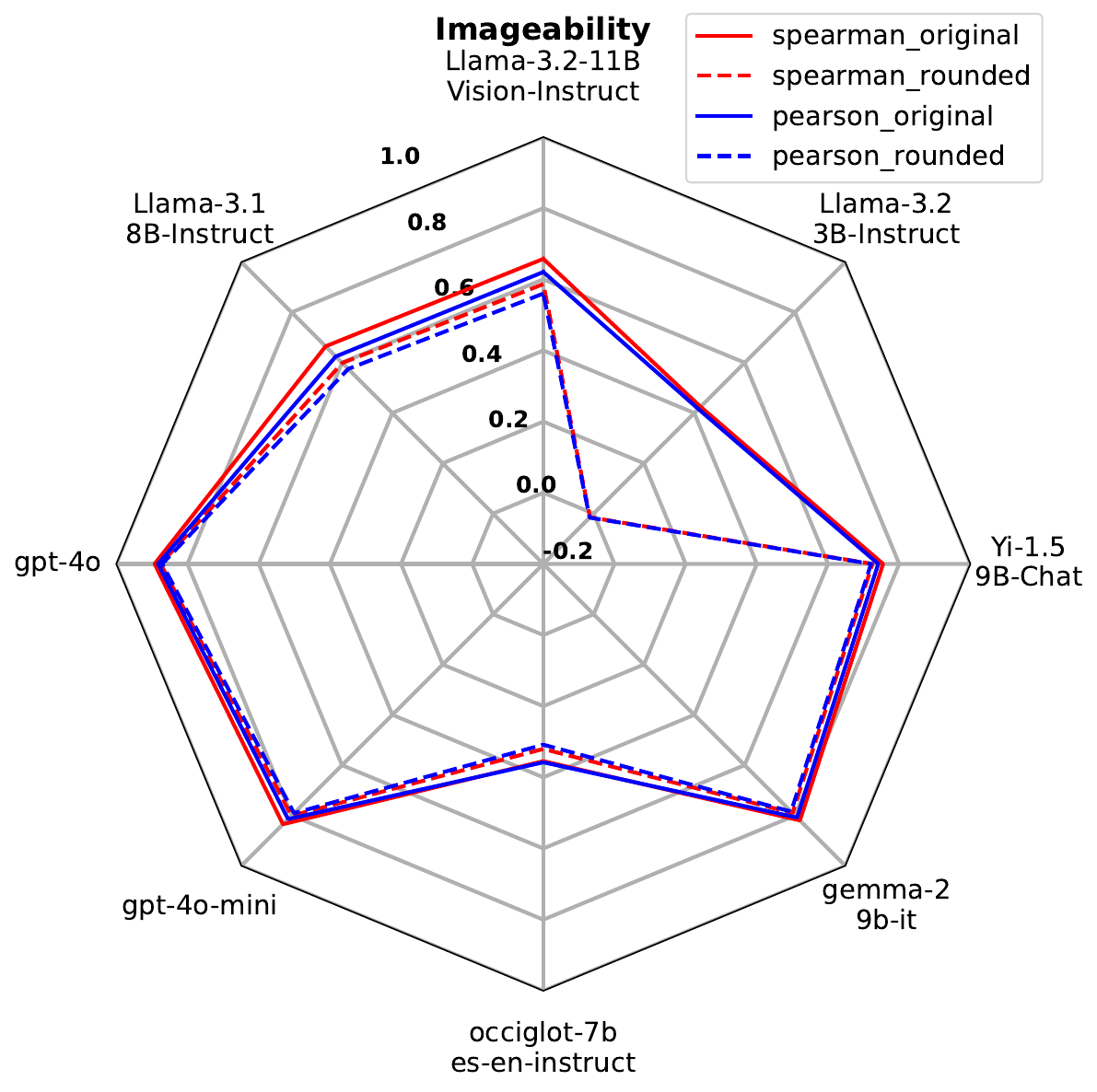}
  
  \caption{Pearson and Spearman correlation coefficients (on original and rounded values) for the Glasgow norms features: \textit{Arousal}, \textit{Valence}, \textit{Concreteness} and \textit{Imageability}.}
  \label{fig:GlasgowRadar1}
\end{figure*}

\subsection{Glasgow norms}

\begin{figure*}[t!]
  \centering
  \includegraphics[width=0.47\textwidth]{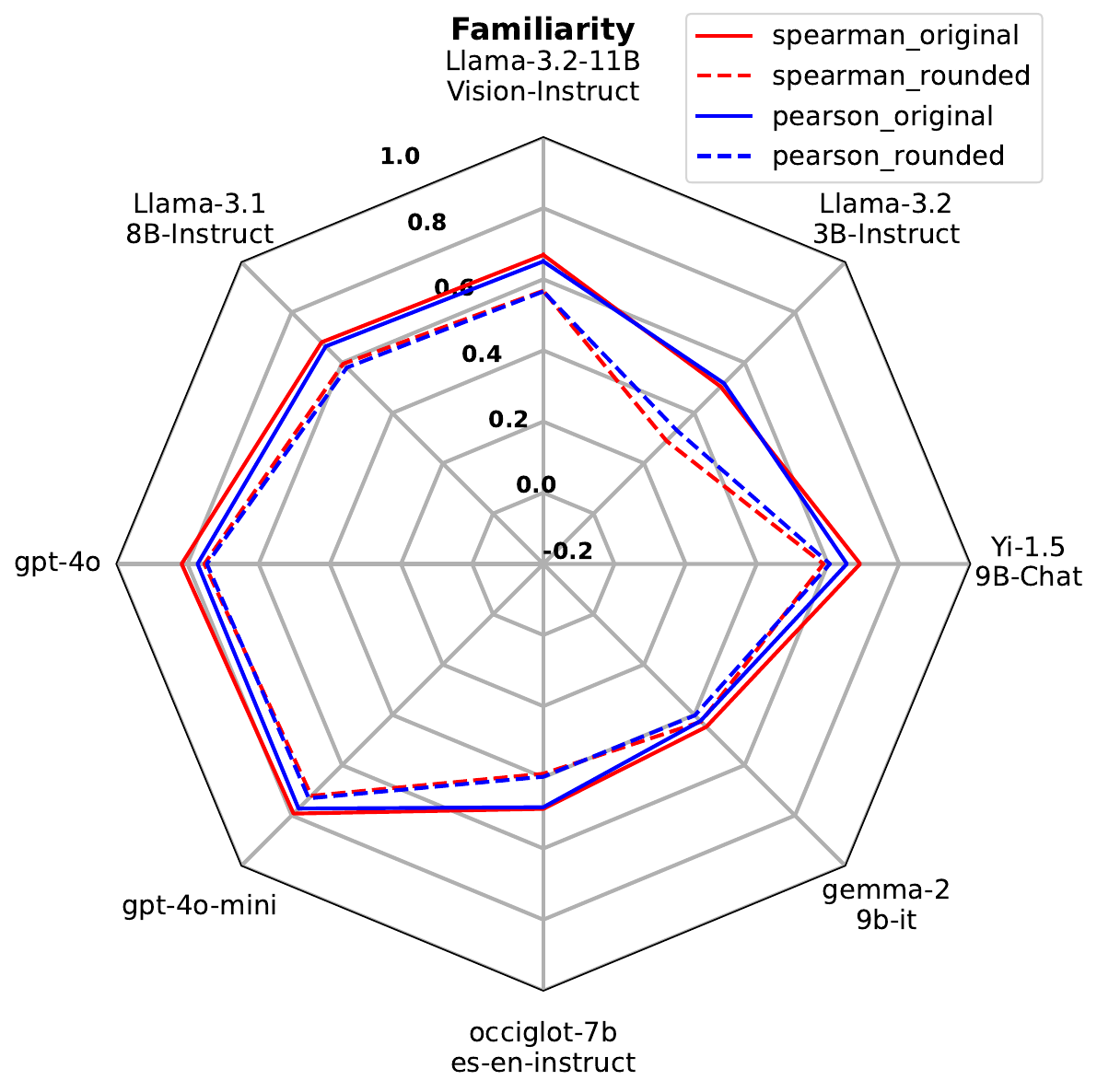}
  \hfill
  \includegraphics[width=0.47\textwidth]{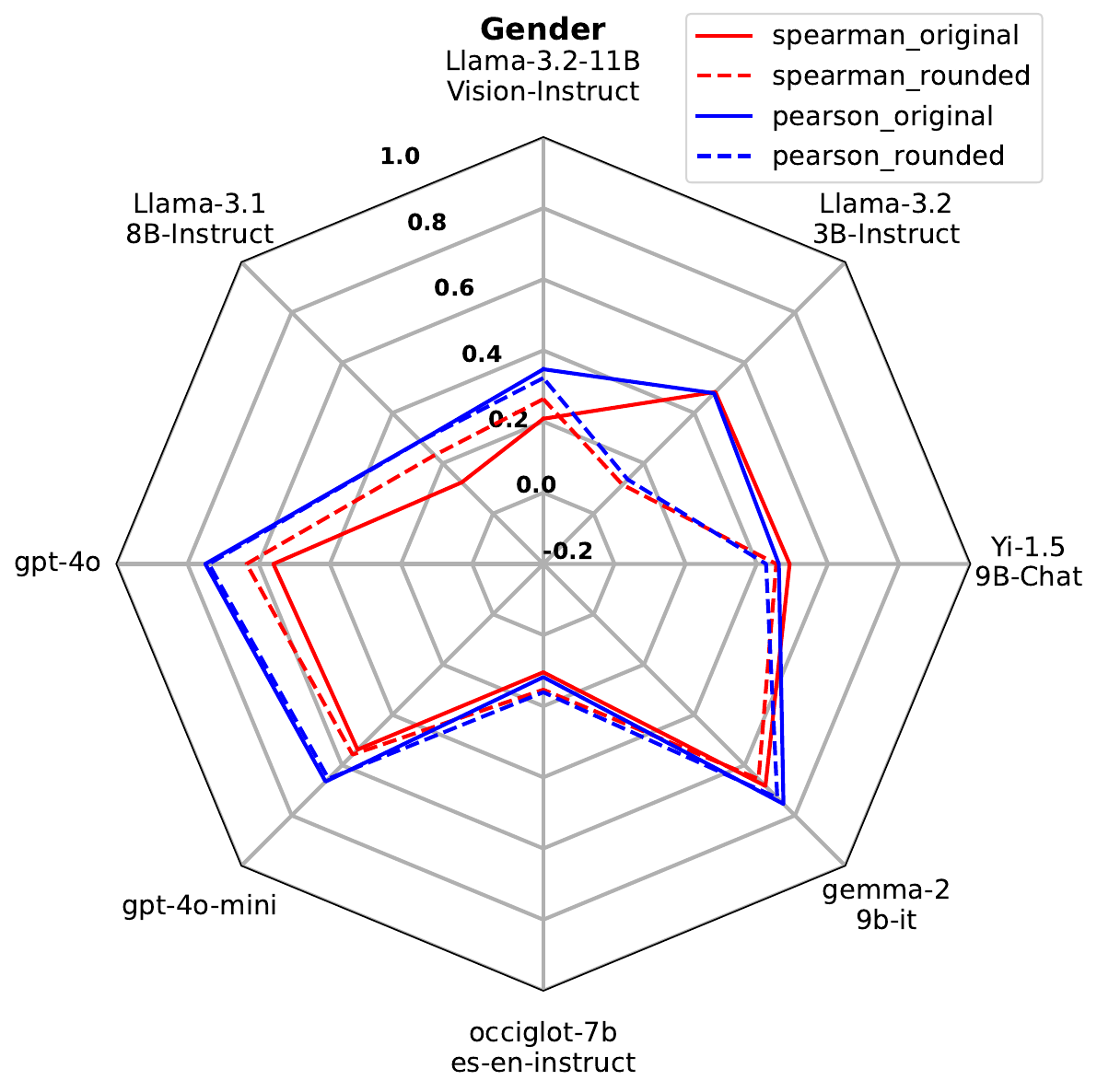}
  \hfill
  \includegraphics[width=0.47\textwidth]{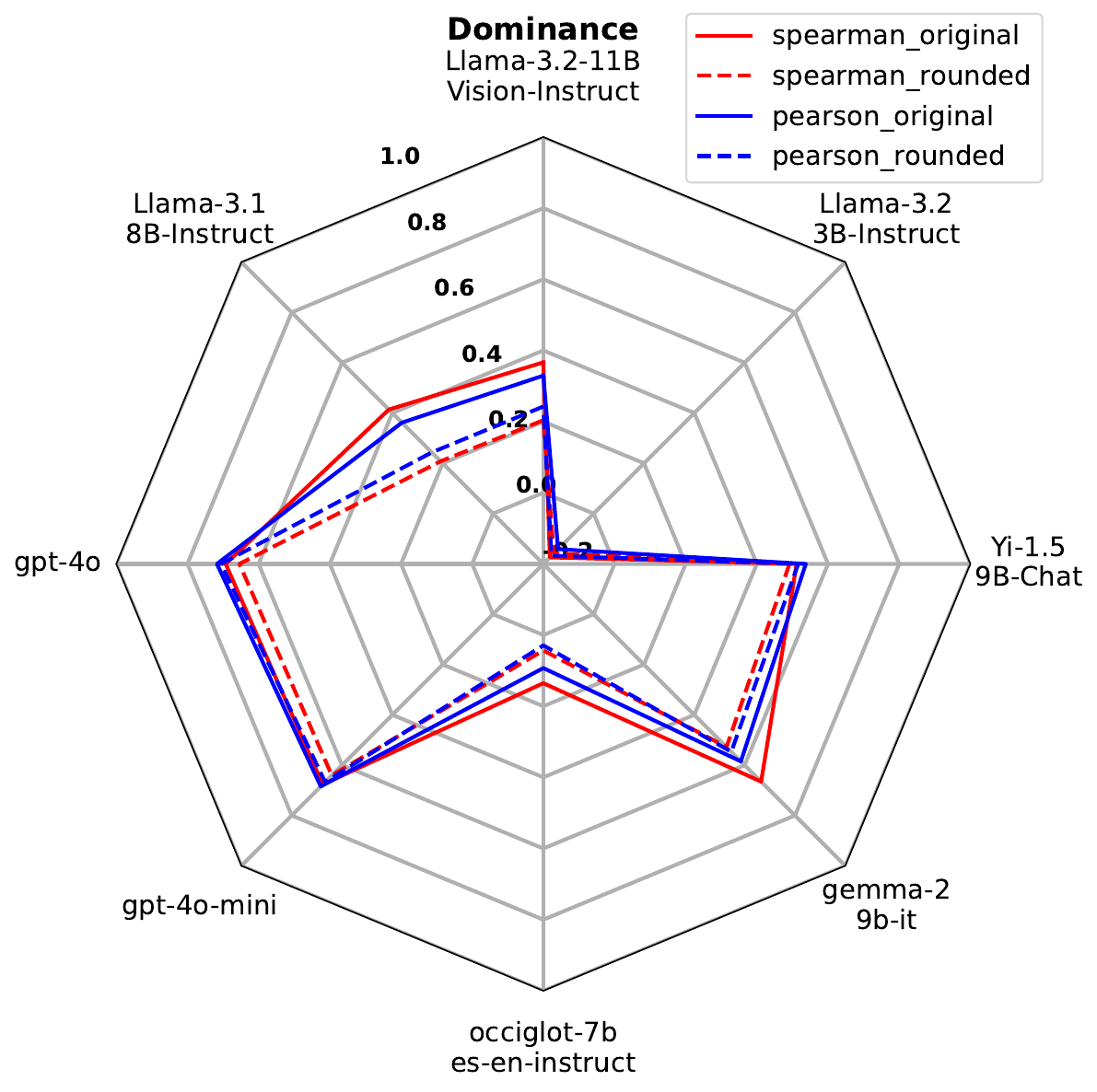}
  \hfill
  
  \caption{Pearson and Spearman correlation coefficients (on original and rounded values) for the Glasgow norms features \textit{Familiarity}, \textit{Gender} and \textit{Dominance}.}
  \label{fig:GlasgowRadar2}
\end{figure*}

The Pearson and Spearman correlation coefficients (both original and rounded) between human and LLM ratings are shown in Figures \ref{fig:GlasgowRadar1} and \ref{fig:GlasgowRadar2} for the seven word features: arousal, valence, concreteness, familiarity, imageability, gender and dominance. Each plot shows the correlation coefficients for a given feature in all models evaluated. It can be seen that alignment is better in general for arousal, valence, concreteness, imageability and familiarity and worse for gender, and dominance. The models with better alignment across all the features are GPT-4o and GPT-4o-mini but other smaller models also have good correlation for some features, for example Gemma-2-9B for gender. Looking at the different correlation coefficients, they generally agree well with a few exceptions. For example, the differences among the coefficients tend to be greater for Llama-3.2-3B.   

In an ideal scenario, the coefficients should be in the 0.8 to 1.0 range (i.e., the outer segment of the web). So, there is room for improvement in the alignment of most models with the features in the Glasgow norms. This confirms the potential of these norms for LLM alignment evaluation.

\textcolor{black}{Two examples of words that get different ratings by humans are \textit{bicycle} and \textit{bid} with 6.81 and 3.42 respectively for concreteness. Instead, Llama-3.2-3B produces similar ratings with values of 4.73 and 4.50 while GPT-4o gets even more extreme values than humans with 7 and 2.96. This shows the differences between models when evaluating the norms.}

\subsection{Lancaster norms}

\begin{figure*}
  \centering
  \includegraphics[width=0.48\textwidth]{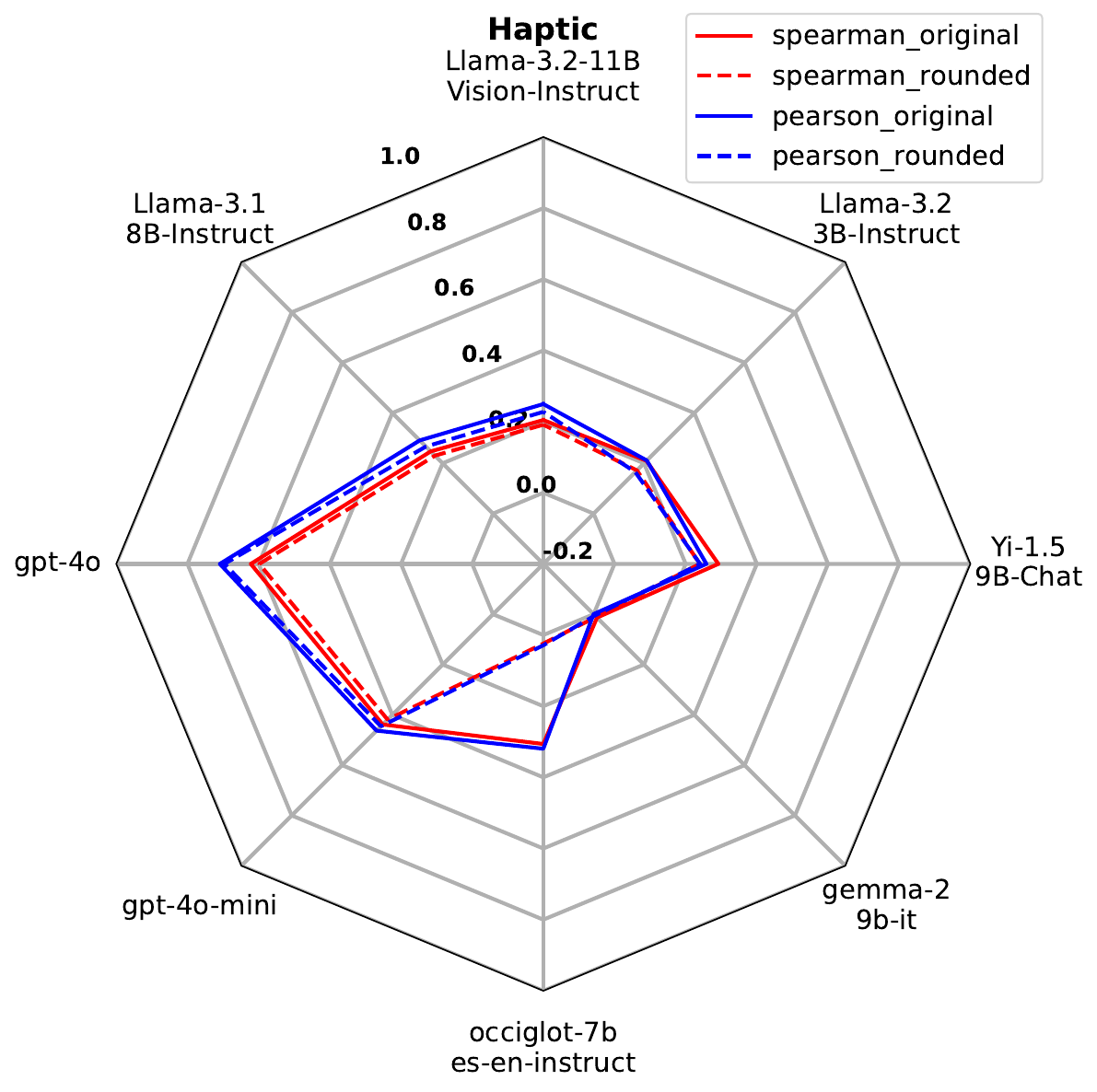}
  \hfill
  \includegraphics[width=0.48\textwidth]{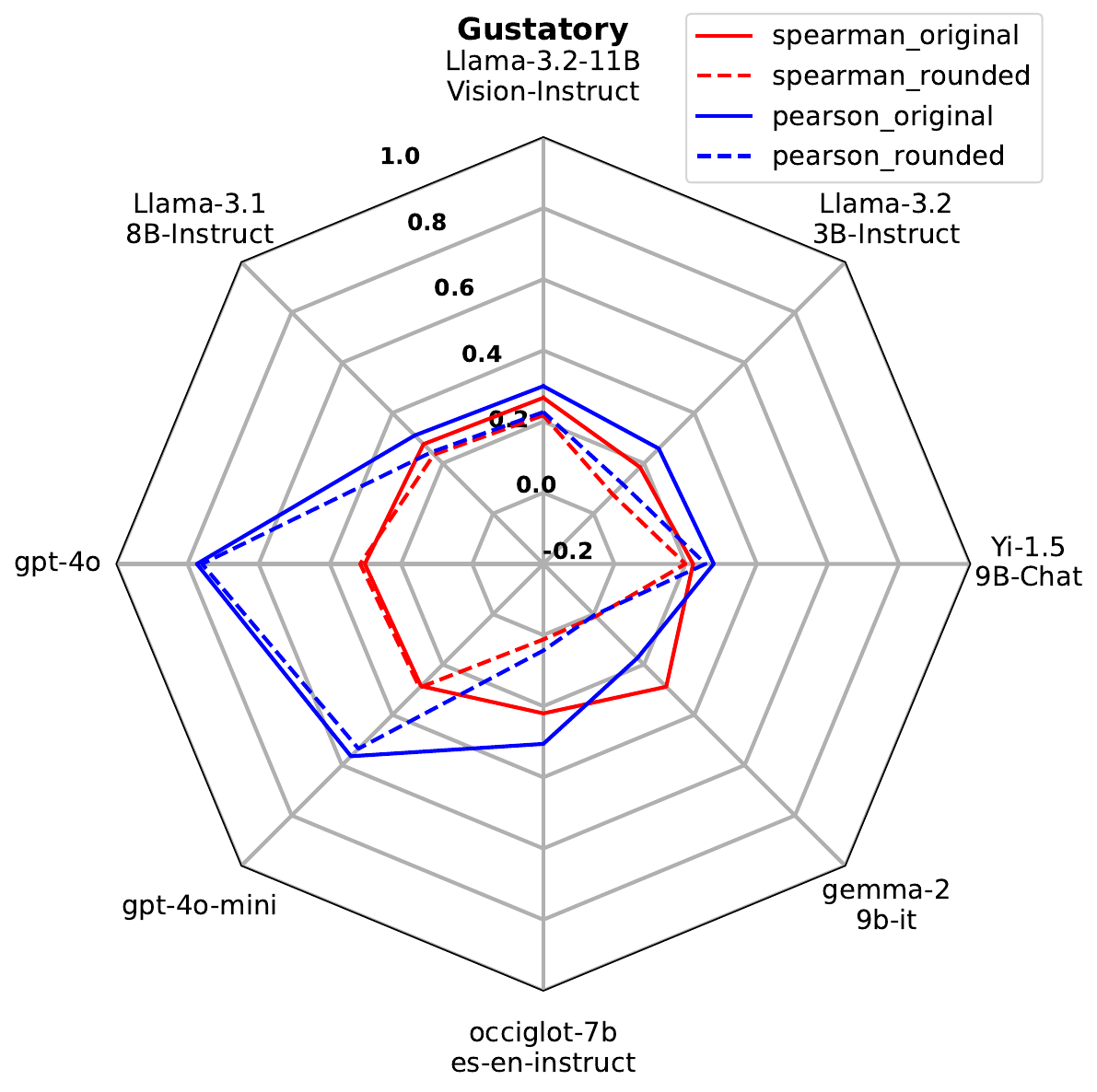}
  \vfill
  \includegraphics[width=0.48\textwidth]{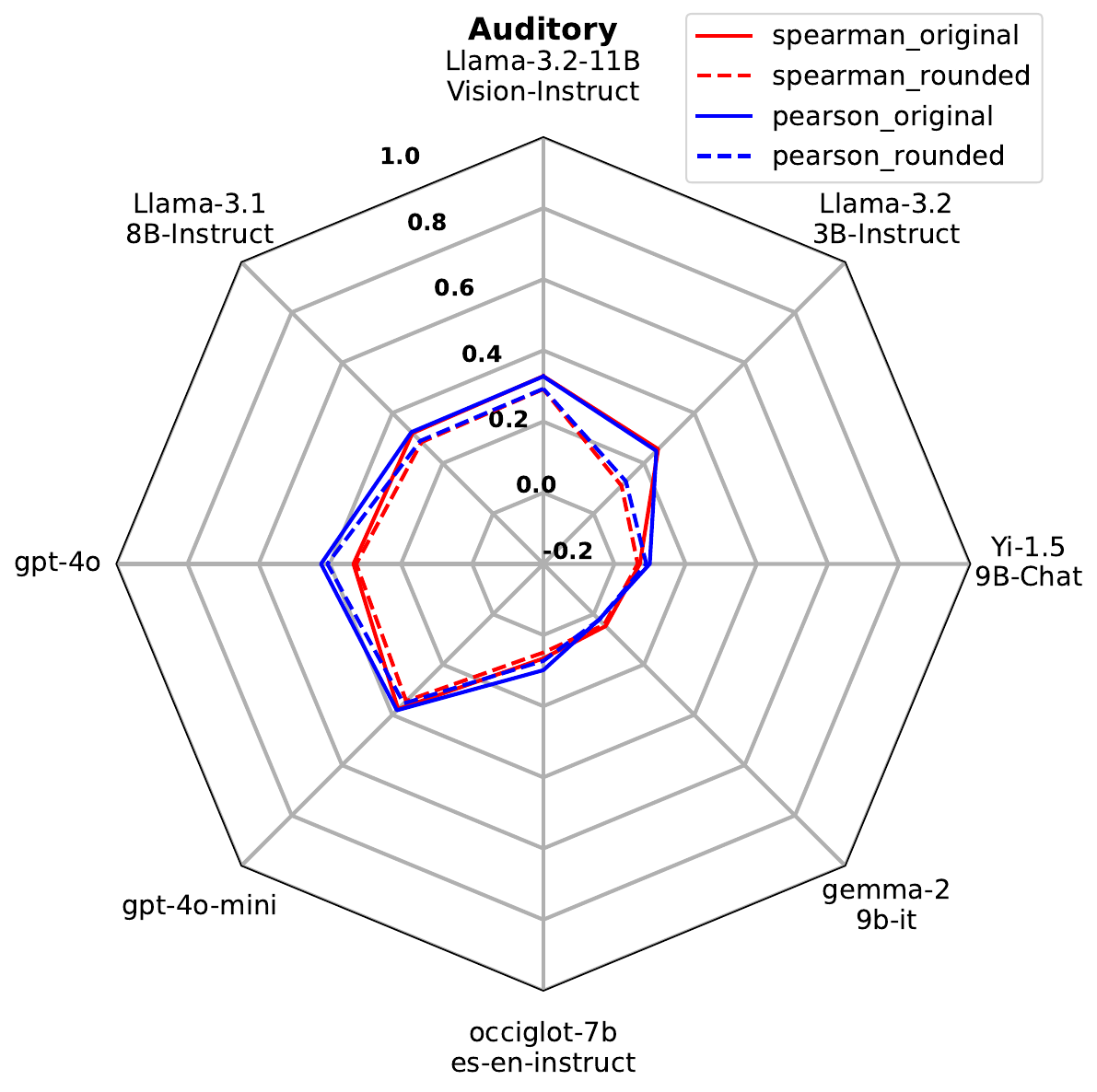}
  \hfill
  \includegraphics[width=0.48\textwidth]{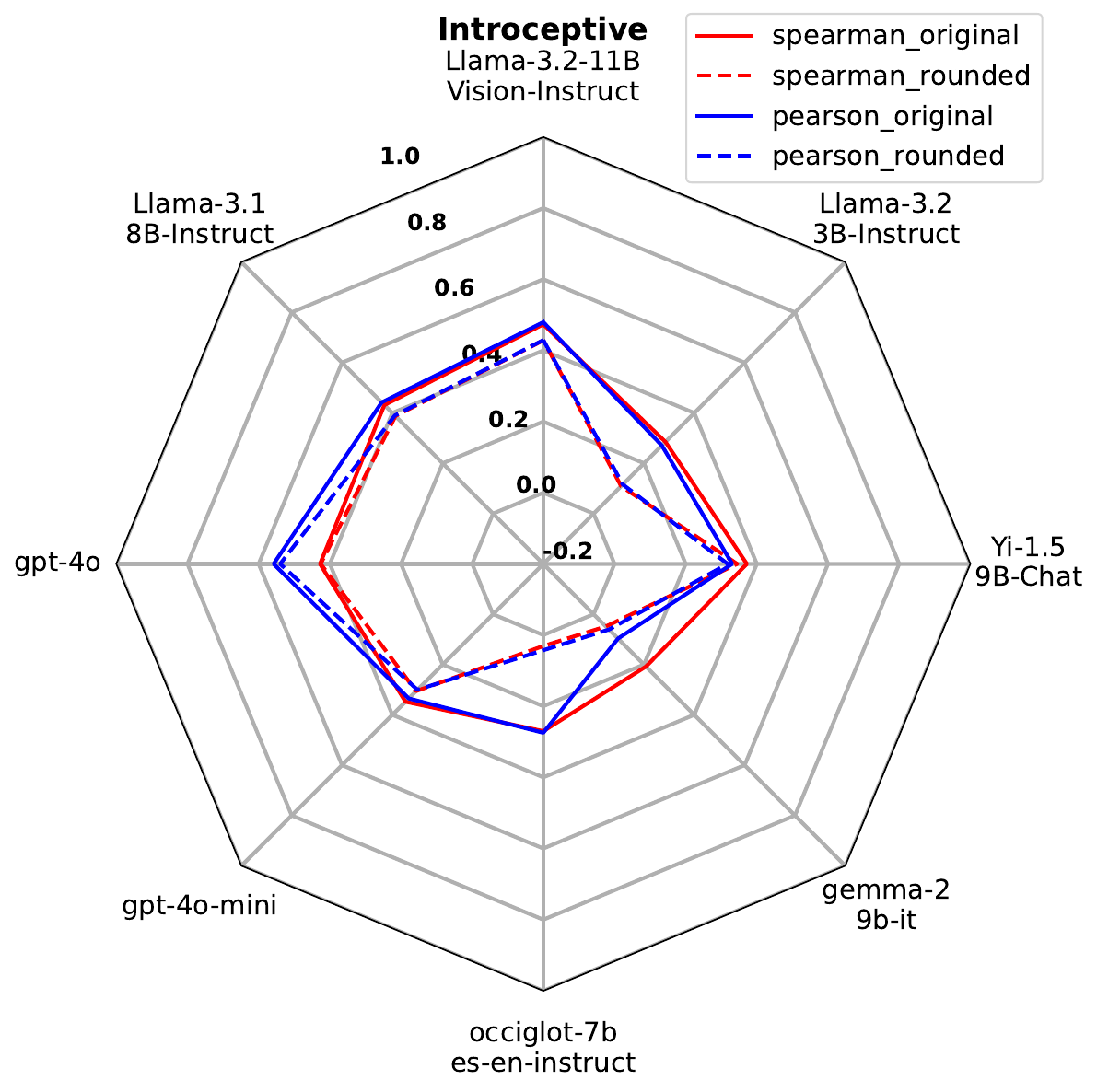}
  \vfill
  \includegraphics[width=0.48\textwidth]{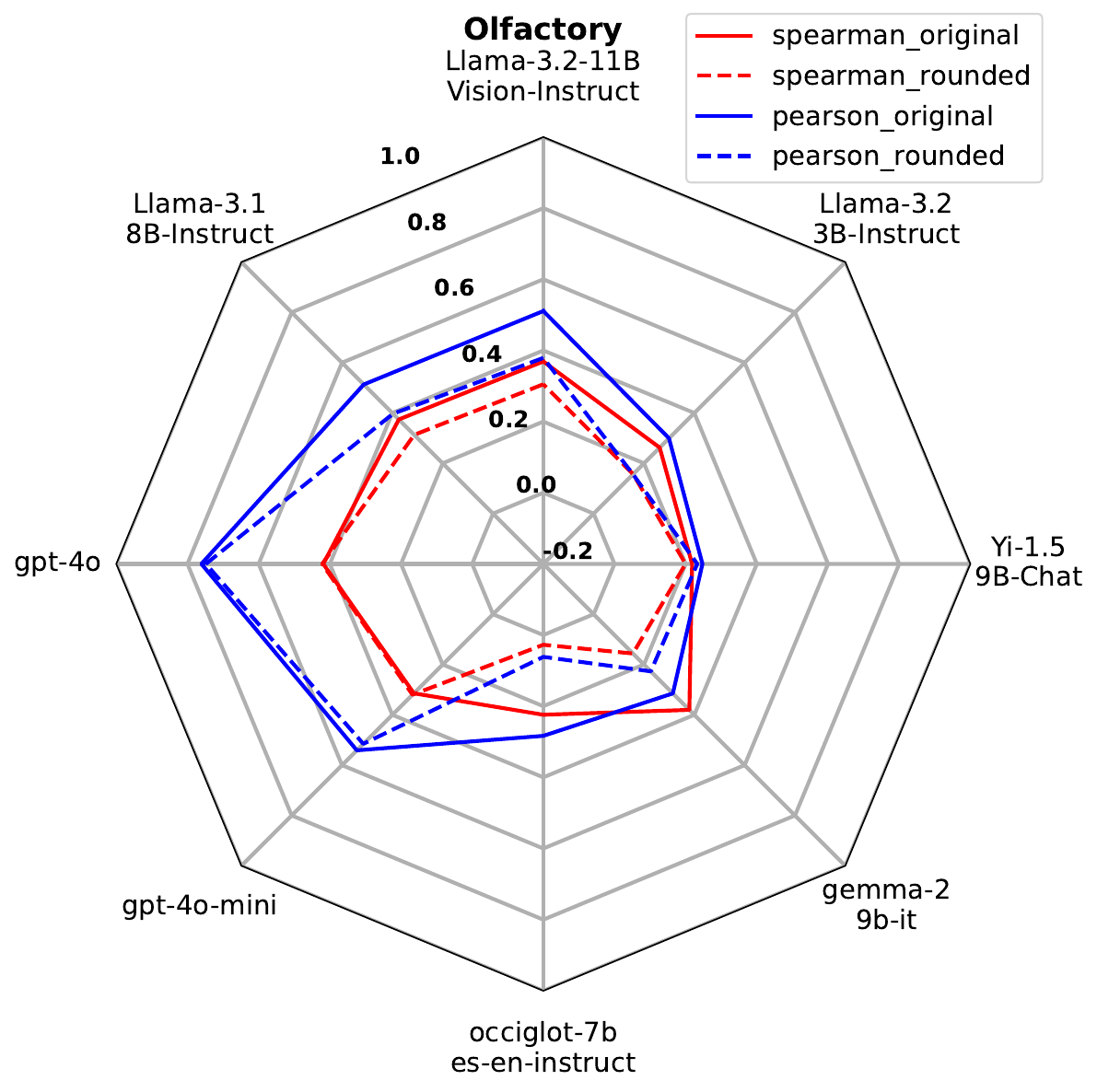}
  \hfill
  \includegraphics[width=0.48\textwidth]{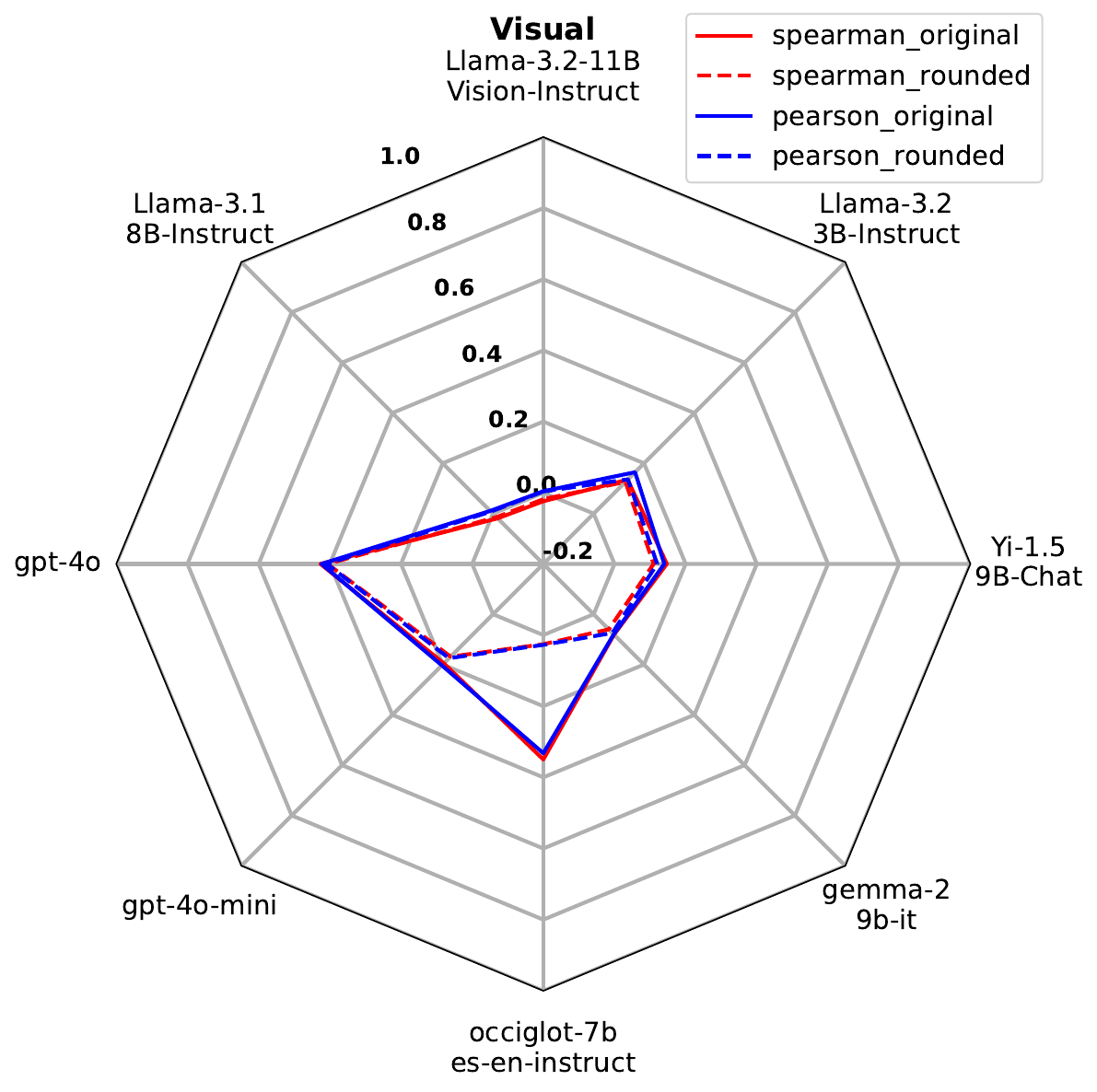}
  
  \caption{Pearson and Spearman correlation coefficients (on original and rounded values) for the Lancaster norms features}
  \label{fig:LancasterRadar}
\end{figure*}

The Pearson and Spearman correlation coefficients (both original and rounded) between human and LLM ratings are shown in Figure \ref{fig:LancasterRadar}. Compared to the results of the Glasgow norms, the correlations are significantly lower, which means that the models are less aligned with humans when it comes to relating words to senses. This may be partially due to the models being trained only with text, as opposed to the additional sensory information available to humans. The best performing model is again GPT-4o but now with much lower correlation values. Comparing among features, olfactory has slightly better results, but still with low correlation coefficients. Multimodality does not seem to help achieve better alignment with the visual feature as multimodal models
(LLama3.2-11B, GPT-4o and GPT-4o-mini) do not have better results than the rest.

The agreement between Pearson and Spearman correlation coefficients is generally good, but not for the gustatory and olfactory ratings. These are the two dimensions with the most skewed distributions (many values at the low end). For these dimensions, the Pearson coefficient (given extra weight to the observations with high values) does considerably better than the Spearman correlation (giving extra weight to differences at the low end of the scale). 

\textcolor{black}{An example of this low correlation is the word \textit{Lemon} with a human rating of 4.45 for gustatory, for which Gemma-2-9B produces a rating of 0.01 although it is a common word directly related to gustatory experience. Instead, GPT-4o produces a rating of 4.49 almost the same as the mean human ratings.}

Considering that correlations would ideally be in the 0.8 to 1.0 range, the current results are very poor and efforts can be made to find out what improves alignment, showing the interest of using the norms for	 LLM evaluation.

\section{Discussion}
\label{sec:discussion}

The results presented in the previous sections show that it is possible to use existing psycholinguistic norms to evaluate the alignment of LLMs with humans on different aspects of word meaning. The methodology proposed is in line with existing LLM evaluation techniques and can be automated, allowing testing at scale. The results also show that the alignment of LLMs is currently limited to a few word norms and only for a few models. Therefore, there is ample room for improvement that future LLMs should address. The alignment is \textcolor{black}{in most cases} worse for perceptual norms, in line with cognitive science results which show that perceptual information is not completely captured by text but also by embodied cognition \cite{barsalou2008grounded,borghi2024language}.

The use of psycholinguistic norms for evaluation has the additional advantage that it provides valuable insights on how to improve LLMs. Now that we know that LLMs lack alignment on perceptual features and that this is probably linked to their lack of embodied cognition, we can start looking into how to train LLMs to acquire that knowledge. We can try generating synthetic text that covers that knowledge and using it in the post-training phase of the LLMs. We can also explore whether multimodal models have the same limitations, for example, for norms related to vision. We will then be able to use the benchmarks to assess the progress made in model alignment when those modifications are introduced. This would promote the participation of the psycholinguistic community in LLM research. 

In fact, more broadly, psycholinguistics can contribute not only to the evaluation of LLMs but also to the understanding of their inner workings and explainability. Psycholinguistics has studied how humans learn and process language for decades, developing theories and experiments to understand our mental processes. In this context, LLMs can be seen as another type of subject to study for which existing knowledge can be reused.

For some models and features we obtained big differences between the Pearson and the Spearman correlations coefficients. To some extent, this is a nuisance as it is unclear which one to rely on. On the other hand, the difference is also informative. Higher Pearson coefficients indicate that observations outside the bulk of the distribution have the desired properties (i.e., the LLM outliers agree with the human outliers). Higher Spearman correlations indicate that small differences around the mode of the distribution align between LLMs and humans. We recommend always computing both correlation coefficients to avoid drawing wrong conclusions (e.g., about quality differences between LLMs in leader boards). Most of the time, rounding to the nearest integer did not make much difference. If it does, this indicates that much of the correlation is due to alignment between models and humans that are unlikely to have psychological significance because they are too small to be noticed by people (e.g., differences between Likert values of 1.01 and 1.02). It is good to check for this possibility if a considerable difference is observed between the Pearson and the Spearman correlation.

\section{Conclusion}
\label{sec:conclusion}

This paper proposes the use of psycholinguistic word norms for the evaluation of human and LLM alignment. The initial results using thirteen word norms covering different aspects of word meaning indicate that current LLMs have limited alignment with humans, \textcolor{black}{and more so} for norms that are related to sensory experiences. This can be linked to the LLMs’ lack of embodied cognition present in humans. The study and results show not only the potential of psycholinguistic word norms for evaluating LLM alignment but also for analyzing the results through the lens of existing psycholinguistic theories.

The methodology, metrics, datasets, and models used in our initial evaluation can be used and extended to define a comprehensive benchmark which can be included in leaderboards as part of the standard LLM evaluation process. This will foster research to improve LLMs’ alignment and the understanding of how models learn and process.

\section*{Limitations}

The initial study on the use of psycholinguistic word norms for LLM evaluation presented in this paper has several limitations. The first is that only two datasets were used and all norms are in English. Additional datasets, norms, and languages should be included to have a comprehensive benchmark similar to those used for task performance evaluation \cite{BIGMeasuring}. Similarly, the number of LLMs evaluated can be extended, ideally including most LLMs in existing leaderboards \cite{open-llm-leaderboard-v2}. The metrics used for evaluation have been taken from psycholinguistic studies, but further analysis is needed to see whether better	metrics can be found for the evaluation of LLM alignment.
 
This work is just an initial step in using psycholinguistic norms to evaluate LLMs. To make this a reality, many additional steps are needed. The first would be to conduct additional evaluations that cover more psycholinguistic datasets and norms, as well as more LLMs. The results of an extensive evaluation could then be used to propose a comprehensive benchmark for assessing LLM alignment, similar to what has been done with language understanding and other tasks \cite{MMLU}. In addition to defining a benchmark, work is needed to explore the metrics used to quantify alignment; the correlation coefficients used in our evaluation are again just a first attempt to measure alignment. Another important consideration is that alignment has to be evaluated not only in English. Therefore, benchmarks in other languages also have to be developed leveraging multilingual word norms to avoid the problems introduced by translating tests, which in the case of word norms could be significant \cite{plaza2024spanish}.


\section*{Acknowledgments}

The work of UPM was supported by the Agencia Estatal de Investigación (AEI) (doi:10.13039/501100011033) under Grants FUN4DATE (PID2022-136684OB-C22) and SMARTY (PCI2024-153434) and by the European Commission through the Chips Act Joint Undertaking project SMARTY (Grant 101140087).

\bibliography{acl_latex}




\end{document}